\documentclass[conference]{IEEEtran}
\usepackage{bm}
\usepackage{times}
\usepackage{xcolor}
\usepackage[numbers]{natbib}
\usepackage{multicol}
\usepackage{amsmath}
\usepackage{amssymb}
\usepackage{graphicx}
\usepackage[bookmarks=true]{hyperref}
\usepackage{float}
\usepackage{booktabs}
\usepackage{graphicx}
\usepackage{placeins}
\usepackage{afterpage}
\usepackage[hypcap=true]{caption}

\begin{document}

\title{Morpheus: A Neural-driven Animatronic Face with Hybrid Actuation and Diverse Emotion Control
}

\newcommand\blfootnote[1]{%
  \begingroup
  \renewcommand\thefootnote{}\footnote{#1}%
  \addtocounter{footnote}{-1}%
  \endgroup
}
\author{Zongzheng Zhang$^{*1, 2}$, Jiawen Yang$^{*1}$, Ziqiao Peng$^{1}$, \\Meng Yang$^{4}$, Jianzhu Ma$^{1}$,  Lin Cheng$^{5}$, Huazhe Xu$^{3}$, Hang Zhao$^{3}$, Hao Zhao\textdagger$^{1, 2}$ \vspace{0.2cm}\\
 $^1$ Institute for AI Industry Research (AIR), Tsinghua University  $^2$ Beijing Academy of Artificial Intelligence (BAAI) \\
 $^3$ Institute for Interdisciplinary Information Sciences(IIIS), Tsinghua University \\
 $^4$ MGI Tech, Shenzhen, China  $^5$ Beihang University \\
 $^*$ Equal contribution \textdagger Corresponding author. Email: zhaohao@air.tsinghua.edu.cn
}



%

\makeatletter
\let\@oldmaketitle\@maketitle
\renewcommand{\@maketitle}{\@oldmaketitle%
    \centering
    \begingroup
    \setcounter{figure}{0}
    \captionsetup{type=figure}
    \includegraphics[width=1.0\linewidth]{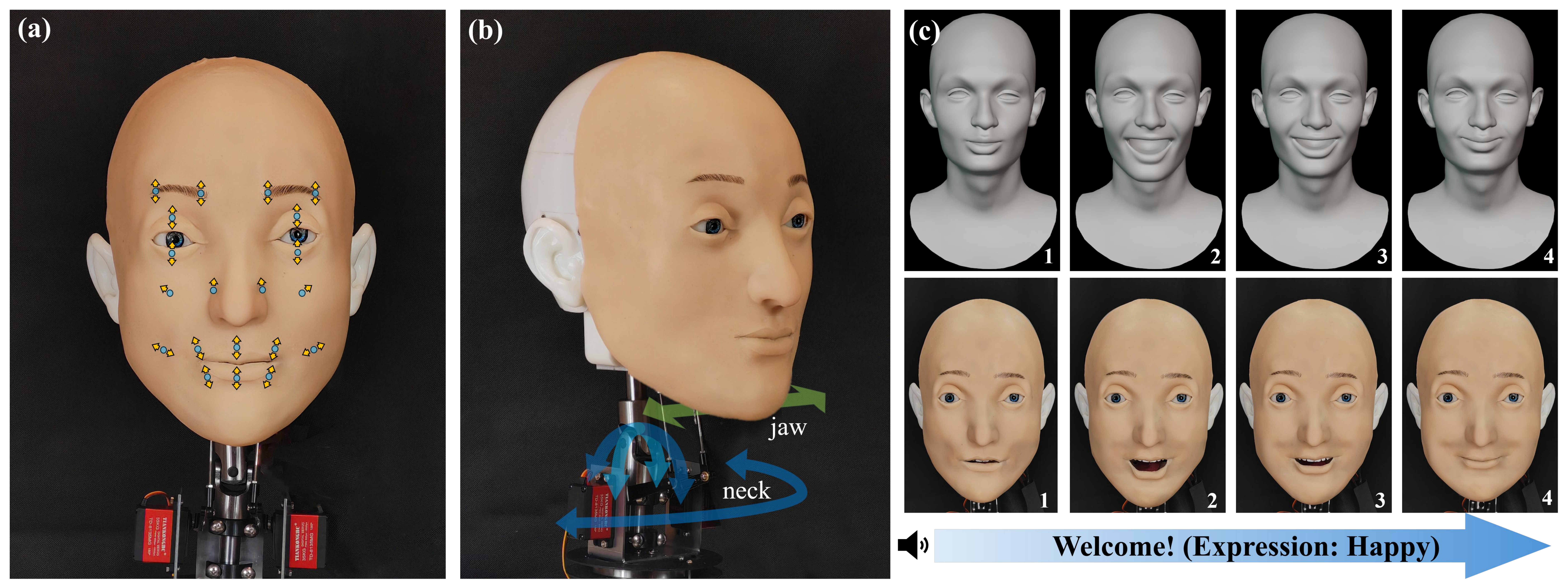}
    \captionof{figure}{\small \textbf{\textit{Morpheus}: an animatronic face with diverse expressions.} (a) Front view. \textcolor{blue}{Blue} markers indicate the attachment points between the underlying mechanical structure and the soft skin, while \textcolor{yellow}{yellow} arrows denote the directions of movement. (b) Side view highlighting motion capabilities. \textcolor{blue}{Blue} arrows indicate the three-axis neck movement: nodding, shaking, and rotation. The \textcolor{green}{green} arrow illustrates the jaw's ability for horizontal movement in addition to typical opening and closing motions, enabling more diverse expressions. (c) Four sequential frames of the "happy" expression as the animatronic face says "welcome!" The first row illustrates the virtual expressions generated by our algorithm rendered in Blender, while the second row displays the corresponding real-world expressions reproduced by the animatronic face.}
    \vspace{-10pt}
    \label{fig:teaser}
    \endgroup

}
\makeatother

\maketitle

\begin{abstract}

Previous animatronic faces struggle to express emotions effectively due to hardware and software limitations. On the hardware side, earlier approaches either use rigid-driven mechanisms, which provide precise control but are difficult to design within constrained spaces, or tendon-driven mechanisms, which are more space-efficient but challenging to control. In contrast, we propose a hybrid actuation approach that combines the best of both worlds. The eyes and mouth—key areas for emotional expression—are controlled using rigid mechanisms for precise movement, while the nose and cheek, which convey subtle facial microexpressions, are driven by strings. This design allows us to build a compact yet versatile hardware platform capable of expressing a wide range of emotions. On the algorithmic side, our method introduces a self-modeling network that maps motor actions to facial landmarks, allowing us to automatically establish the relationship between blendshape coefficients for different facial expressions and the corresponding motor control signals through gradient backpropagation. We then train a neural network to map speech input to corresponding blendshape controls. With our method, we can generate distinct emotional expressions such as happiness, fear, disgust, and anger, from any given sentence, each with nuanced, emotion-specific control signals—a feature that has not been demonstrated in earlier systems. We release the hardware design and code at \href{https://github.com/ZZongzheng0918/Morpheus-Hardware}{https://github.com/ZZongzheng0918/Morpheus-Hardware} and \href{https://github.com/ZZongzheng0918/Morpheus-Software}{https://github.com/ZZongzheng0918/Morpheus-Software}.

\end{abstract}
 
\section{Introduction}

Imagine a museum guide robot explaining exhibits to children \cite{bennewitz2005towards, kuno2007museum}. Despite having flexible limbs and responsive question-answering capabilities, these robots may still lack a sense of engagement and immersion. The absence of nuanced emotional expressions through facial features prevents them from evoking a sense of closeness. In human-robot interaction, a mechanical face capable of dynamic and expressive movements is crucial for bridging the emotional gap, fostering trust and empathy through a tangible, three-dimensional medium for emotional communication. This makes it indispensable in applications such as healthcare, education, and entertainment.

However, existing animatronic face systems encounter challenges in effectively conveying emotions due to constraints in both hardware and software design. In terms of hardware design, conventional systems rely on either fully rigid-driven mechanisms or fully tendon-driven actuation. While rigid-driven systems \cite{brooks1998cog, kobayashi2000study, lee2008development, hashimoto2008dynamic, lutkebohle2010bielefeld, ishihara2011realistic, yan2024facial, hu2024human} offer precise control over movement trajectories, they struggle to handle subtle facial regions like the nose and cheek due to space constraints, which are crucial for conveying microexpressions. On the other hand, tendon-driven systems \cite{li2024driving} are more space-efficient but fall short when it comes to controlling areas such as the eyebrows and mouth, where larger and more dynamic movements are essential for effective emotional expression. In contrast, \textit{Morpheus}, our hardware design (refer to Figure \ref{fig:teaser}(a), (b)), adopts a hybrid actuation approach, utilizing 33 actuators to combine the strengths of both rigid and tendon-driven mechanisms. This design enables the creation of more dynamic and diverse facial expressions within a compact space, effectively addressing the limitations of previous systems.

Fig.~\ref{fig:teaser}(c) illustrates \textit{Morpheus} generating a happy expression while saying "Welcome!" in both the virtual and real domains. The mouth shape, facial dynamics, and overall expression closely align with the intended emotional content of the speech. Precise control is essential for breathing life into the mechanical structure, enabling the creation of complex and diverse facial expressions. Traditional expression control methods which rely on predefined control programs~\cite{oh2006design, berns2006control, mazzei2012hefes, asheber2016humanoid}, result in limited, static expressions and fail to produce smooth, continuous transitions between different emotions. Currently, learning-based methods mainly focus on replicating expressions observed from human faces via sensors~\cite{yan2024facial}, such as cameras~\cite{chen2021smile}, or predicting facial movements for a specific future frame~\cite{hu2024human}. For instance, ~\citet{li2024driving} explores robot facial expression generation from speech. However, it still falls short in capturing the facial expressions that naturally occur during speech, resulting in mechanical, emotionless detached playback. Current methods remain far from achieving diverse, expressive, and contextually appropriate speech-driven robotic facial expressions. 

On the algorithmic side, digital human researchers have developed blendshape bases that can generate a wide array of facial expressions. However, the challenge remains in effectively translating these blendshapes into motor control signals for an animatronic face. The key issue is identifying a control basis that aligns with the blendshapes, as animatronic faces, when integrated with skin and tendons, become too complex to model entirely for control purposes. The complexity stems from the non-linear and often unpredictable relationship between motor actions and facial expressions, which is exacerbated when mechanical systems interact with soft tissues. To tackle this, we adopt a self-modeling approach.
We train a network using random pairs of motor control signals and corresponding animatronic face landmarks to establish a relationship between them. Once trained, this network is fixed. By applying backpropagation, we can then calculate the necessary motor control signals to achieve specific blendshapes. This method provides a systematic and precise way to control the animatronic face's motors to replicate the desired expressions.

We then develop a neural network to translate speech into a set of blendshape coefficients. When integrated with the motor control signal coefficient determined by our self-modeling network, it enables our animatronic face to produce a broad spectrum of emotional expressions in response to spoken language. We can now generate a diverse range of emotional responses, from joy to fear, disgust to anger, each meticulously aligned with the emotional nuances of the spoken words. This capability to produce nuanced and contextually appropriate expressions in real-time represents a significant leap forward in animatronic technology.

We summarize our three contributions below:
\begin{itemize}

\item We propose a hybrid-actuated robotic face hardware platform that combines the advantages of both rigid and tendon-driven mechanisms, allowing for precise control of key areas for emotional expression while maintaining space efficiency.
\item We introduce a self-modeling network that maps motor actions to facial landmarks, enabling the automatic establishment of the relationship between blendshape coefficients and motor control signals through gradient backpropagation.
\item We train a neural network to map speech input to corresponding blendshape controls, enabling the generation of distinct emotional expressions with nuanced, emotion-specific control signals.
\end{itemize}
\section{Related Works}

\begin{figure*}
    \centering
    \includegraphics[width=1.0\linewidth]{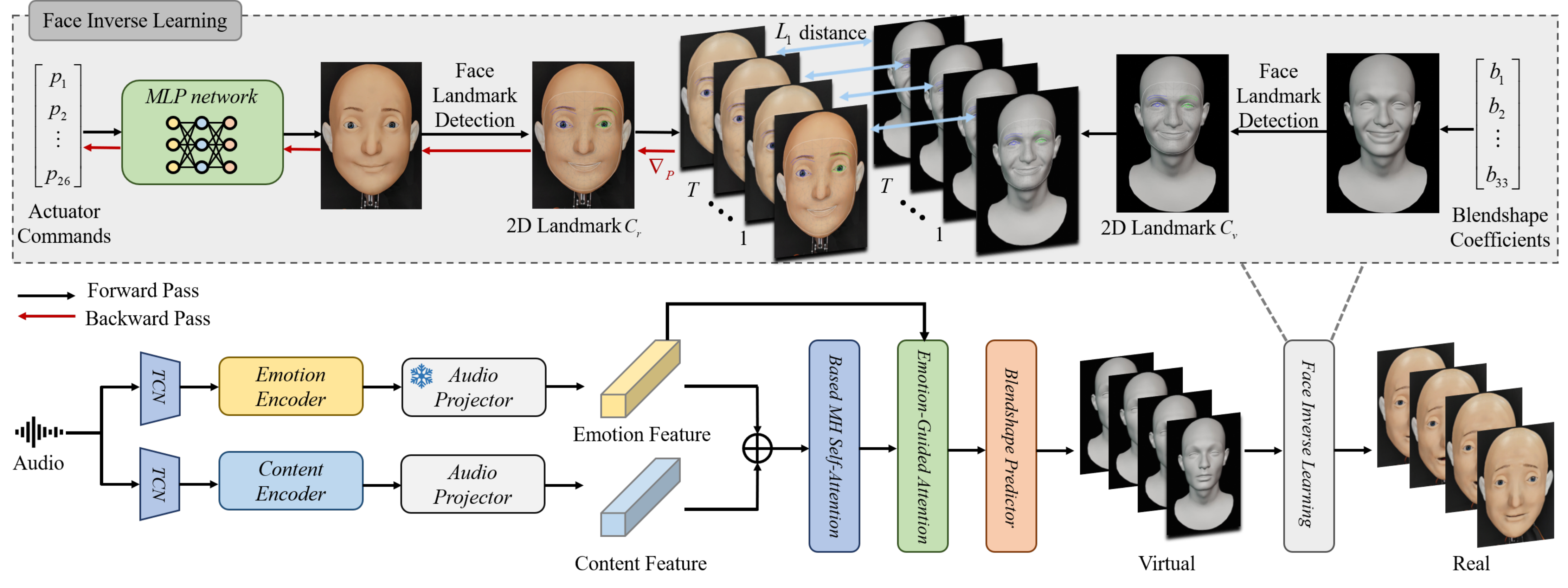}
    \caption{\small \textbf{Overview of the Mechanism for Generating Realistic Speech-driven Facial Expressions.} We propose a method that disentangles content and emotion from expressive speech and generates blendshape coefficients using a transformer-based decoder with emotion-guided attention. These coefficients are then mapped to actuator commands through a face inverse learning module, ultimately producing realistic facial expressions that match the emotional tone of the speech.}
    \vspace{-10pt}
    \label{fig:overview}
\end{figure*}

\subsection{Audio-Driven Facial Animation}

Facial animation and emotional expression synthesis have gained significant attention in both computer graphics and human-robot interaction, with applications ranging from virtual avatars to robotic systems. Early work on facial animation, such as the Viseme-Phoneme model~\cite{zhou2018visemenet}, focuses on matching facial movements to phonemes for lip-syncing but fails to capture emotional variations in speech. This limitation makes the generated expressions less realistic beyond basic lip movements.

Recent advancements in deep learning~\cite{cudeiro2019capture, fan2022faceformer,peng2023selftalk, richard2021meshtalk}, particularly using Transformers, have improved facial animation by predicting facial movements directly from audio features. These methods capture more nuanced aspects of speech, including the dynamics of facial expressions in response to both speech content and emotional tone. However, many of these models still prioritize the content of speech, with less focus on the emotional dimension.

In response to this gap, models like Ji's~\cite{ji2021audio} emotion disentangling approach and EmoTalk~\cite{peng2023emotalk} have started integrating emotional features in facial animation. Ji's model separates emotional and content representations in speech, allowing for more expressive and dynamic facial movements. EmoTalk further extends this concept to 3D face animation tasks. Our work builds on these ideas by offering a more integrated approach to emotion and content disentangling, improving the stability of the process and enhancing the synchronization between audio and facial movements.

\subsection{Mechanical Face}

Over the years, various humanoid robots have been developed to enhance human-robot interaction through facial expressions. Early works by \citet{kobayashi1993study}, \cite{kobayashi2000study}, \cite{kobayashi2003realization}, \citet{hashimoto2006development}, \cite{hashimoto2008dynamic}, \citet{berns2006control} utilize flexible micro-actuators and pneumatic systems to replicate human-like facial expressions, improving emotional communication  \cite{itoh2006mechanical, oh2006design}. Later advancements, such as the use of rope-driven mechanisms \cite{weiguo2004development}, provide more realistic and compact facial movements. \citet{brooks1998cog} introduces a humanoid robot head with 21 degrees of freedom (DOFs) to integrate facial expressions with social behaviors. Additionally, bio-inspired actuators, including electro-active polymers (EAP) \cite{hanson2002identity}, are introduced to enhance expression flexibility. Humanoid robots, HRP-4C \cite{kaneko2009cybernetic, nakaoka2009creating}, Janet and Thomas\cite{lin2009realization} and Eva \cite{faraj2021facially},  have emphasized full-body integration with advanced facial expression capabilities.  \citet{allison2009design} proposes a unique muscle actuation system for socially assistive robots, utilizing fewer actuators to mimic muscle activity and create natural facial expressions. Furthermore, the Flob robot \cite{lutkebohle2010bielefeld} integrates high-resolution sensors for improved social interaction through facial expressions. The Affetto child robot \cite{ishihara2011realistic} uses detailed facial design to study caregiver-child attachment, while the EveR-2 platform \cite{lee2008development} emphasizes an expressive humanoid design for emotional communication in human-robot interactions.

More recently, \citet{li2024driving} has proposed a tendon-driven approach for animatronic robots, using linear blend skinning (LBS) for speech-synchronized facial movements. \citet{yan2024facial} develops a humanoid robot head with a rigid linkage drive system to generate six basic facial expressions. \citet{hu2024human} presents Emo, a rigid-driven humanoid robot platform with 26 actuators and enhances facial expressivity through anticipatory modeling, enabling real-time co-expression of facial expressions with humans. In this paper, We adopt a hybrid drive system combining tendon and rigid actuators, utilizing 33 actuators to achieve diverse facial expression control.

\subsection{Inverse Expression Learning}

Pioneering works by \citet{breazeal2003emotion},  \citet{berns2006control},  \citet{nakaoka2009creating} and \citet{yan2014survey} rely on pre-programmed systems to generate facial expressions, which limits their flexibility in real-time emotional responses. Due to the complexity and non-linearity of robot face, expression recognition   \cite{gu2017local, ishihara2018identification, liu2017facial, song2009image}, and learning-based method are crucial. Recently, researchers have made significant progress by integrating facial expression control techniques, such as HEFES system \cite{mazzei2012hefes}, forward and inverse kinematics models \cite{ren2016automatic},  genetic algorithms \cite{habib2014learning, hyung2019optimizing}, CNN-LSTM-based emotion classification system \cite{liu2019emotion}, visual mimicry learning \cite{chen2021smile, hu2024human}, Facial Action Coding System \cite{sato2022android}, Bayesian optimization \cite{yang2022optimizing}, MAP-Elites algorithm \cite{tang2023automatic} and visual self modeling \cite{chen2022fully}. 

Since 2023, Gaussian Splatting \cite{kerbl20233d} has introduced efficient differentiable rasterization, and has been extended to model dynamics \cite{wu20244d, liu2024differentiable}. Given the motion retargeting approach proposed in \cite{liu2024differentiable}, we backpropagate gradients for optimizing actuator commands, thereby establishing a connection between virtual expressions and real-world actuator commands.

\section{Method}
\label{sec:method}

\begin{figure*}
    \centering
    \includegraphics[width=1.0\linewidth]{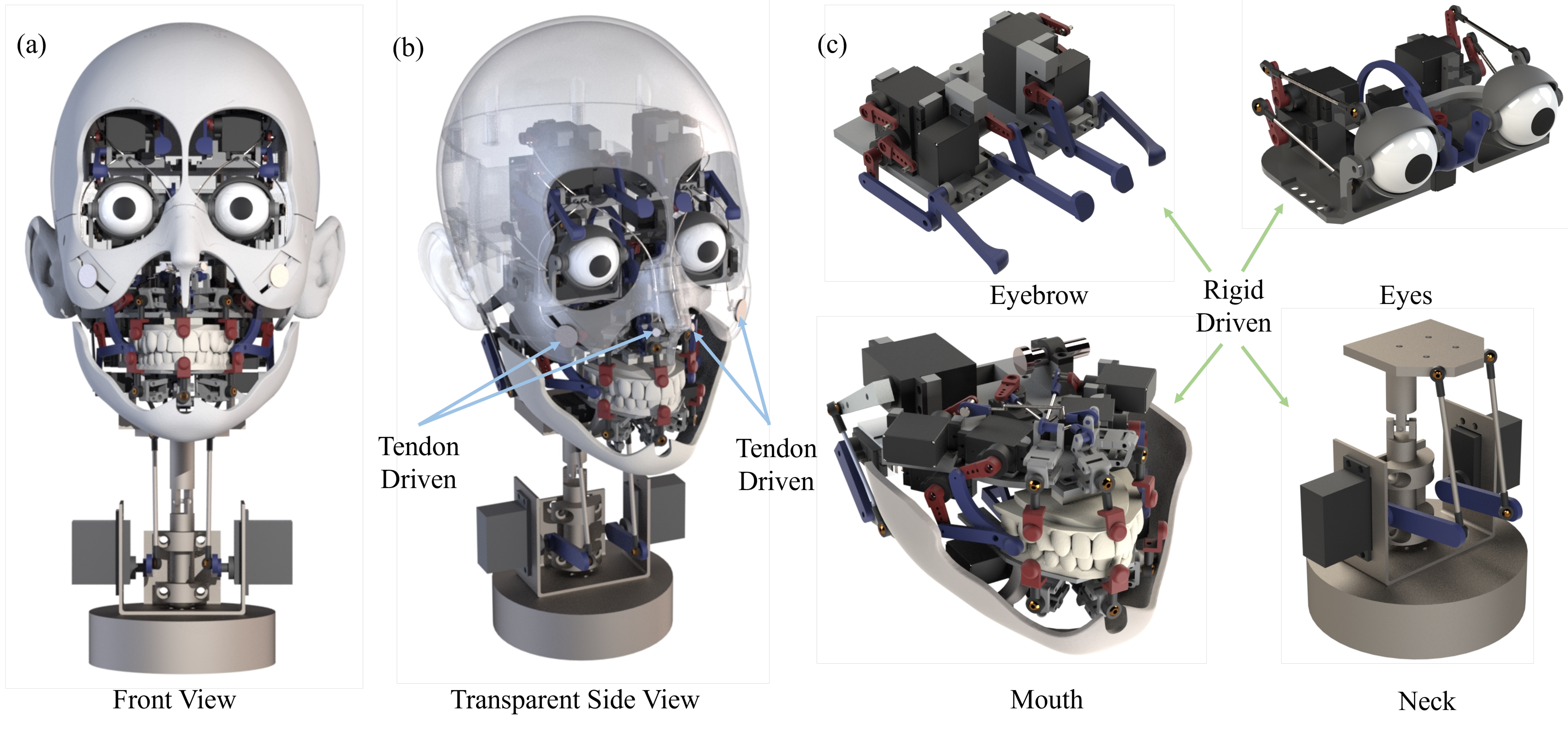}
    \caption{\small \textbf{Hardware Design of Our Mechanical Face Platform.} (a) Front view of the mechanical skeletal structure. (b) Transparent side view of the mechanical structure, highlighting tendon-driven movements in the nose and cheek areas (indicated by \textcolor{blue}{blue} arrows). (c) Four rigid-driven modules (indicated by \textcolor{green}{green} arrows), including the eyebrow module (4 actuators), eye module (6 actuators), mouth module (16 actuators), and neck module (3 actuators), work together to form the complete mechanical face platform.}
    \vspace{-10pt}
    \label{fig:mechanical_face}
\end{figure*}

This section begins with an overview of the architecture (Sec.~\ref{section3.1}) integrating the components of our system. Next, we detail the algorithmic framework (Sec.~\ref{section3.2}) for transforming speech inputs into virtual emotional expressions and the hardware design (Sec.~\ref{section3.3}) of the robotic face. We then introduce the face inverse learning module (Sec.~\ref{section3.4}), which maps virtual expressions to motor commands. Finally, the overall loss function (Sec.~\ref{section3.5}) for optimizing the system is outlined.

\subsection{Architecture Overview}
\label{section3.1}

Our speech-driven robotic face system with varying expressions consists of three main components: the \textit{Emotional Facial Animation}  module, the \textit{Mechanical Face Platform}, and the \textit{Face Inverse Learning} module. As shown in the Fig.~\ref{fig:overview}, when a speech input with emotional tone  $\boldsymbol{A}_{1:T}$  is fed into the Emotional Facial Animation module, the model generates a sequence of blendshape coefficients $\boldsymbol{B}_{1:T}$, which define the virtual facial expressions predicted for each frame. These coefficients are then passed to the Face Inverse Learning module, which outputs actuator control commands  $\boldsymbol{P}_{1:T}$ that closely match the virtual facial expressions. Finally, the generated facial expression sequence is implemented on the Mechanical Face Platform, synchronized with the speech input.

\subsection{Emotional Facial Animation}
\label{section3.2}

In this part, our goal is to create expressive 3D facial animations that not only reflect the emotional tone of the spoken content but also offer precise control over the emotional intensity and personal expression style, which are crucial for enhancing the realism of mechanical faces. The model takes a sequence of speech snippets \( \boldsymbol{A}_{1:T} = (\boldsymbol{a}_1, \dots, \boldsymbol{a}_T) \), where each \( \boldsymbol{a}_t \in \mathbb{R}^D \) is a frame of audio, and produces corresponding facial blendshape coefficients \( \boldsymbol{B}_{1:T} = (\boldsymbol{b}_1, \dots, \boldsymbol{b}_T) \in \mathbb{R}^{33} \), which define the facial movements at each time step. These blendshape coefficients control the movement of a mechanical face, ensuring that it expresses the emotional content of the speech in a natural and dynamic manner.

To achieve high-quality facial animations, we tackle the challenge of disentangling the speech content from the emotional tone. Our solution involves an emotion disentangling encoder, which separates the content and emotion features of the speech signal. We use pre-trained audio models fine-tuned for extracting distinct content and emotion features. The encoder is trained on pseudo-training pairs with varying emotional tones and content, ensuring that it learns to independently represent speech content and emotional expression. This disentanglement enables the model to accurately map the emotional aspects of speech to facial movements, independent of the specific content of the speech.

Since speech samples with the same content but different emotions often have different speech rates, directly using these samples without alignment may lead to temporal mismatches in facial animations. To address this, we apply Dynamic Time Warping (DTW)~\cite{berndt1994using} to align the feature sequences. Specifically, we use intermediate features extracted from a pre-trained wav2vec model~\cite{baevski2020wav2vec}, which provides robust and expressive audio representations. Given two wav2vec feature sequences \( \bm{S}_a \) and \( \bm{S}_b \) of the same content but different lengths, DTW calculates a set of index coordinate pairs \( \{ (i, j), \dots \} \) by dynamic programming to align \( \bm{S}_a[i] \) and \( \bm{S}_b[j] \). The optimal alignment path minimizes the cumulative distance between corresponding features:

\begin{equation}
\begin{aligned}
\min \sum_{(i,j) \in P} d(\bm{S}_a[i], \bm{S}_b[j]),
\end{aligned}
\end{equation}
where \( d \) is the Euclidean distance cost and \( P \) is the alignment path. The path constraint ensures that valid steps are \( (i+1, j) \), \( (i, j+1) \), or \( (i+1, j+1) \), enforcing forward progression in at least one of the sequences at each step. By aligning the speech samples in this manner, we ensure that temporal inconsistencies in the emotional and content features are minimized, leading to smoother and more synchronized facial animations.

Once the content and emotion features are extracted and temporally aligned, they are fed into a feature fusion decoder to generate the 3D facial blendshape coefficients. The decoder integrates these features to produce realistic facial motions. We adopt a transformer-based architecture for the decoder, which effectively captures the temporal dependencies and dynamic nature of facial movements. Positional encoding is used to model the timing of facial actions, while a multi-head self-attention mechanism helps the model focus on local, temporal patterns crucial for generating smooth, coherent facial animations. An emotion-guided attention mechanism is also introduced to prioritize emotional features, ensuring that the generated facial expressions are strongly influenced by the emotional tone of the speech.

Finally, the fused features are passed through a feed-forward layer and decoded into 33 blendshape coefficients using a fully connected layer. These coefficients control the mechanical face's movements, producing expressive and emotionally resonant facial animations. This approach enables the generation of highly customized facial motions for mechanical faces, with users able to adjust both emotional intensity and personal style. By combining emotion disentangling, temporal alignment through DTW, and emotion-guided attention, our method provides a powerful tool for creating realistic, dynamic, and emotionally expressive facial animations for mechanical avatars.

\subsection{Hardware Design}
\label{section3.3}

The primary objective of the hardware design (Fig.~\ref{fig:mechanical_face}(a)) is to mimic diverse human expression. The system utilizes a total of 33 actuators, 29 of them are distributed in the rigid-driven part: 4 actuators in the eyebrow module, 6 in the eye module, 16 in the mouth module, and 3 in the neck module (Fig.~\ref{fig:mechanical_face}(c)); the rest 4 of them are distributed in the tendon-driven part controlling the motion of nose and cheek (Fig.~\ref{fig:mechanical_face}(b)). Integrated movement of these modules allows for fundamental and diverse human facial expressions, as well as neck twists and nods. The main structure of the hardware is built through 3D printing. Through the integration of these modules, the proposed hardware design achieves a high degree of realism in mimicking human facial expressions, making it a robust platform for advanced human-robot interaction. For a more detailed demonstration of the mechanical face's functionality, please refer to the \textcolor{red}{Supplementary Material}.

\subsubsection{Module Design and Actuation}

A sufficient number of actuators provides the possibility of diverse expressions. The decomposition of facial expressions is a fundamental prerequisite for achieving such versatility.  

In this work, the movement of the \textbf{eyebrows} is divided into the movement of the head of the eyebrows and the peak of the eyebrows. These two motions are relatively defined, so two sets of four-bar mechanisms are employed to realize them. It is worth noting that according to the motion principle of the human eyelid, only the upper eyelid moves when blinking. However, the lower eyelid moves upward due to the contraction of the orbicularis oculi muscle. Consequently, the lower eyelid also needs to be configured with degrees of freedom to achieve realistic motion. 

The \textbf{eye} movement is implemented as a two-degree-of-freedom movement, reflecting the fact that human eye movements are predominantly synchronous. To achieve realism and reduce mechanical complexity, the motions of the two eyeballs are linked, enabling them to rotate simultaneously to any desired position. Realistic eyeballs are equipped with miniature cameras for face detection. This setup fosters natural engagement during conversations, enriching the overall interaction and creating a sense of connection with the external environment. 

In the \textbf{mouth} module, the lips are discretized into eight control points: two at the corners of the mouth and three at each of the upper and lower lips. The movement of the two corner points is driven by a planar five-bar mechanism, which has a wide range of movement. This mechanism allows the platform to replicate facial movements associated with joy, such as the upward movement of the mouth corners, as well as expressions of sadness or anger, characterized by downward movements. Six points on the lips control the opening and closing of the lips. Additionally, the jaw, beyond its normal opening and closing motion, is capable of lateral translation using a rack-and-pinion mechanism. This level of discretization provides sufficient precision to emulate the complex dynamics of lip movements during human speech.

\subsubsection{Hybrid Actuation Control}

The simulated face mechanism uses a hybrid actuation system that combines tendon-driven (Fig.~\ref{fig:mechanical_face}(b)) and rigid-driven (Fig.~\ref{fig:mechanical_face}(c)) components. Specifically, the \textbf{nose} and \textbf{cheek} areas are driven by strings, while the rest of the face is driven by traditional rigid mechanisms. Tendon drives are highly flexible and take up little space, making them particularly suitable for applications such as the human face, where space is limited and the number of degrees of freedom is high. The principle is also similar to that of a real muscle stretching drive. Therefore, it's suitable for the complex movement of the cheek and nose. On the other hand, rigid-driven systems offer greater reliability, load-bearing capacity, and well-defined trajectories. The eyebrow, eye and mouth modules utilize such drives. By combining the flexibility of tendon-driven systems with the stability and precision of rigid-driven mechanisms, this hybrid approach enables a more natural and expressive simulation of facial movements.

\subsection{Face Inverse learning}
\label{section3.4}

The \textit{Face Inverse Learning} module plays a critical role in transforming blendshape coefficients into precise  actuator commands for the mechanical face platform. Due to the complexity of soft skin as a nonlinear system, where the movement state is influenced by factors such as material properties and attachment points, traditional model-based control methods are no longer applicable. In this work, we adopt an inverse learning approach to bridge the gap between virtual facial expressions and the real-world actuator-driven movements.

Excluding the two motors controlling eye movements, three motors for neck movements (their movements are controlled by an additional facial tracking system), and two motors for tongue movements, the remaining 26 motors \(\boldsymbol{P} = (p_1, \dots, p_{26})\)  are responsible for generating facial expressions. As Fig.~\ref{fig:overview}
 shows, we build an MLP network to model diverse facial expressions. Using the face landmark detection module in MediaPipe~\cite{lugaresi2019mediapipe}, we obtain the 2D landmarks $\boldsymbol{C}_r = (c_{r_1}, \dots, c_{r_N}) \in \mathbb{R}^{468 \times 2}$ of the real facial expressions. This allows us to complete the mapping process using the dataset we construct (Sec.~\ref{4.1}). To accelerate the network optimization process and reduce parameters, similarly to the approach used in ~\citet{hu2024human}, we decouple the facial expression into two independent parts: the upper part (controlled by the eyebrow and eye modules) and the lower part (controlled by the mouth module), improving training speed without sacrificing accuracy.

 Similarly, the 2D landmarks $\boldsymbol{C}_v = (c_{v_1}, \dots, c_{r_N}) \in \mathbb{R}^{468 \times 2}$ from virtual faces are also obtained. By calculating the $\ell_1$ distance between these two sets of landmarks, we can back-propagate the gradient $\nabla p$ to the self-modeling face, thereby optimizing the 26-dimensional motor control parameters.
 
\subsection{Loss Function}
\label{section3.5}

To optimize our neural network for generating realistic and expressive facial animations for mechanical faces, we define a comprehensive loss function that incorporates two key components: cross-reconstruction loss, self-reconstruction loss. The overall objective function is a weighted sum of these individual losses, as shown below:

\vspace{-5pt}
\begin{equation}\label{4}
\mathcal{L} = \lambda_1 \mathcal{L}_{\text{cross}} + \lambda_2 \mathcal{L}_{\text{self}},
\end{equation}
where the hyperparameters \( \lambda_1 = 1.0 \), \( \lambda_2 = 1.0 \) are set empirically based on our experiments. Each loss term plays a distinct role in guiding the network toward learning high-quality facial motions and emotional expressions, ensuring smooth temporal dynamics, and improving emotional recognition accuracy. Below, we provide detailed descriptions of each loss component.

\noindent\textbf{Cross-Reconstruction Loss.} This loss component is crucial for disentangling the emotional $e_1, e_2$ and content  features $c_1, c_2$ from the speech signal. The network is trained to reconstruct facial blendshape coefficients based on different combinations of content and emotion. Given audio inputs \( \boldsymbol{A}_{c_1, e_2} \) and \( \boldsymbol{A}_{c_2, e_1} \), the encoder extracts content and emotional features, which are then processed by the decoder to produce new combinations of facial movements. The cross-reconstruction loss measures the discrepancy between the decoder's output and the corresponding ground truth blendshape coefficients, \( \boldsymbol{B}_{c_1, e_1} \) and \( \boldsymbol{B}_{c_2, e_2} \). Mathematically, this is expressed as:

\vspace{-5pt}
\begin{equation}\label{5}
\begin{aligned}
\mathcal{L}_{\text{cross}} = & \, \left\| \boldsymbol{D}\left( \boldsymbol{E}_c\left( \boldsymbol{A}_{c_1, e_2} \right), \boldsymbol{E}_e\left( \boldsymbol{A}_{c_2, e_1} \right) \right) - \boldsymbol{B}_{c_1, e_1} \right\|_2^2 \\
& + \left\| \boldsymbol{D}\left( \boldsymbol{E}_c\left( \boldsymbol{A}_{c_2, e_1} \right), \boldsymbol{E}_e\left( \boldsymbol{A}_{c_1, e_2} \right) \right) - \boldsymbol{B}_{c_2, e_2} \right\|_2^2,
\end{aligned}
\end{equation}
where \( \boldsymbol{D} \) represents the decoder responsible for generating the facial blendshape coefficients, guided by the emotion and content features extracted by the encoders \( \boldsymbol{E}_c \) and \( \boldsymbol{E}_e \).

\noindent\textbf{Self-Reconstruction Loss.} To maintain the quality of facial animation, we also incorporate a self-reconstruction loss, which ensures that the network can accurately reconstruct the original blendshape coefficients from the same content and emotional input. This encourages the network to generate consistent facial motions for each specific speech sample. The self-reconstruction loss is given by:

\vspace{-5pt}
\begin{equation}\label{6}
\mathcal{L}_{\text{self}} = \left\| \boldsymbol{D}\left( \boldsymbol{E}_c\left( \boldsymbol{A}_{c_1, e_2} \right), \boldsymbol{E}_e\left( \boldsymbol{A}_{c_1, e_2} \right) \right) - \boldsymbol{B}_{c_1, e_2} \right\|_2^2.
\end{equation}
This loss helps the model focus on maintaining high fidelity in the generated facial expressions, ensuring that it can faithfully reproduce the target blendshape coefficients from a given input sequence.

In the \textit{Face Inverse Learning} module, the loss function for the MLP is defined as the $\ell_1$ distance $\mathcal{L}_{\mathit{L_1}}$. Specifically, it is calculated as:

\vspace{-5pt}
\begin{equation}\label{7}
    \begin{aligned}
\mathcal{L}_{\mathit{L_1}}(C_r, C_v) = \frac{1}{N} \sum_{i=1}^{N} \| c_{r_i} - c_{v_i} \|_1,
    \end{aligned}
\end{equation}
where $\boldsymbol{C}_r = (c_{r_1}, \dots, c_{r_N})$ and $\boldsymbol{C}_v = (c_{v_1}, \dots, c_{v_N})$ represent the set of 2D landmarks of the real and virtual facial expression, respectively, and \( N \) is the number of 2D landmark points. Minimizing this loss generates the precise actuator commands for real face expressions that closely match the given virtual faces.
 \section{Experiments}

\subsection{Implementation Details}
\label{4.1}

\textbf{Datasets.} To train our \textit{Emotional Facial Animation} module, we use the 3D-ETF dataset~\cite{peng2023emotalk}. This dataset includes multi-emotion facial animation data with a total duration of five hours. Two widely used 2D audio-visual datasets are utilized to construct the 3D-ETF dataset: RAVDESS~\cite{livingstone2018ryerson} and HDTF~\cite{zhang2021flow}. By combining these datasets, the 3D-ETF dataset provides different emotions and high-resolution facial animation data, enabling our model to learn nuanced emotional expressions.

To train the \textit{Face Inverse Learning} module, we construct a dataset for our system. As described in Sec.~\ref{section3.4}, 26 motors control various facial expressions.  To avoid structural interference during the random generation of control commands, we impose constraints on each motor's movement commands beforehand. Since each motor can have multiple possible positions, the number of possible facial expressions that can be generated is exponentially large.

For dataset creation, we position an RGB camera in front of the mechanical face and randomly send motor control instructions. These expressions are recorded by MediaPipe's~\cite{lugaresi2019mediapipe} face landmark detection, capturing the 2D landmark values. This methodology allows us to construct a dataset comprising 5000 samples—five times larger than the dataset created in~\citet{hu2024human}—which is used to train the MLP network. The resulting dataset offers a comprehensive set of training samples that map actuator commands to their corresponding facial expressions, effectively completing the process of modeling the transition from motor control signals to realistic facial movements.

\textbf{Hardware Deployment.} The simulated face mechanism employs TD-8035MG servos in the neck module, with a blocking torque of 35$~\mathrm{kg\cdot cm}$ to carry the platform's weight and the rest of the servos are GUOHUA 9g A0090 micro servos with 4.6$~\mathrm{kg\cdot cm}$ blocking torque. The soft skin is made from zero-degree liquid silicone, molded by a custom mole. The primary control board is an NVIDIA Jetson Xavier, capable of processing audio, images and video data and send control signals to the motors. 

\subsection{Diverse Expressions}

\begin{figure}
    \centering
    \includegraphics[width=1.0\linewidth]{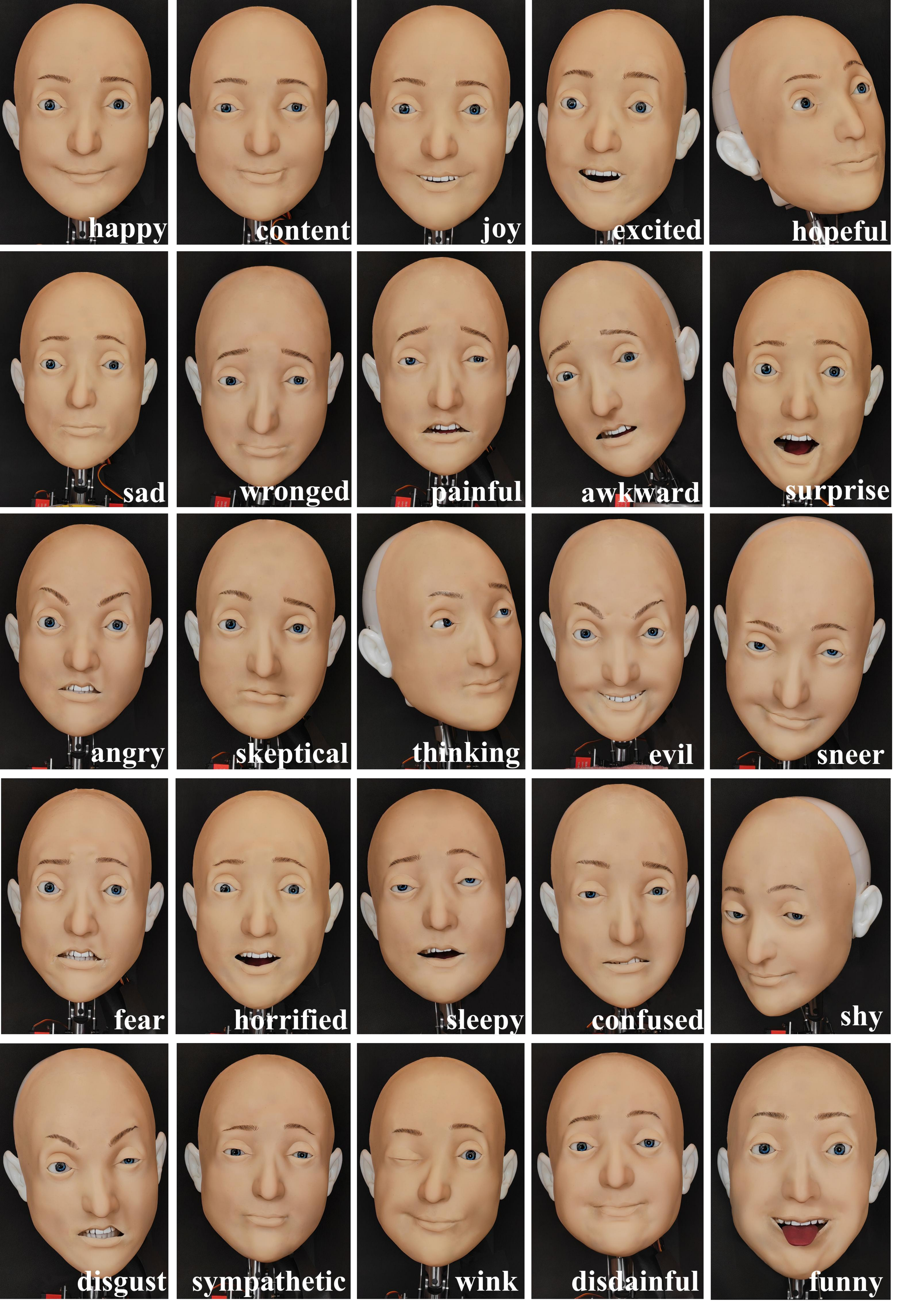}
    \caption{\small \textbf{Expressions generated by the mechanical face platform.} The platform demonstrates 25 emotional expressions, from "happy" to "funny," showcasing its ability to replicate a variety of human facial movements, providing a basis for realistic human-robot interaction.}
    \vspace{-10pt}
    \label{fig:diverse_expressions}
\end{figure}

Fig.~\ref{fig:diverse_expressions} shows 25 different human facial expressions we selected, ranging from basic emotions such as "angry," "happy," "sad," "fear," "disgust," to more nuanced ones like "skeptical," "shy," and "sympathetic." These expressions are generated through precise control of the eyebrow, eye, mouth and neck modules, showcasing the hardware system's ability to produce a wide variety of rich and complex facial movements.  The platform's ability to replicate such a broad range of emotional states demonstrates its versatility 
   and strong potential for realistic human-robot interaction.

\subsection{Quantitative Results}

\begin{figure*}
    \centering
    \includegraphics[width=1.0\linewidth]{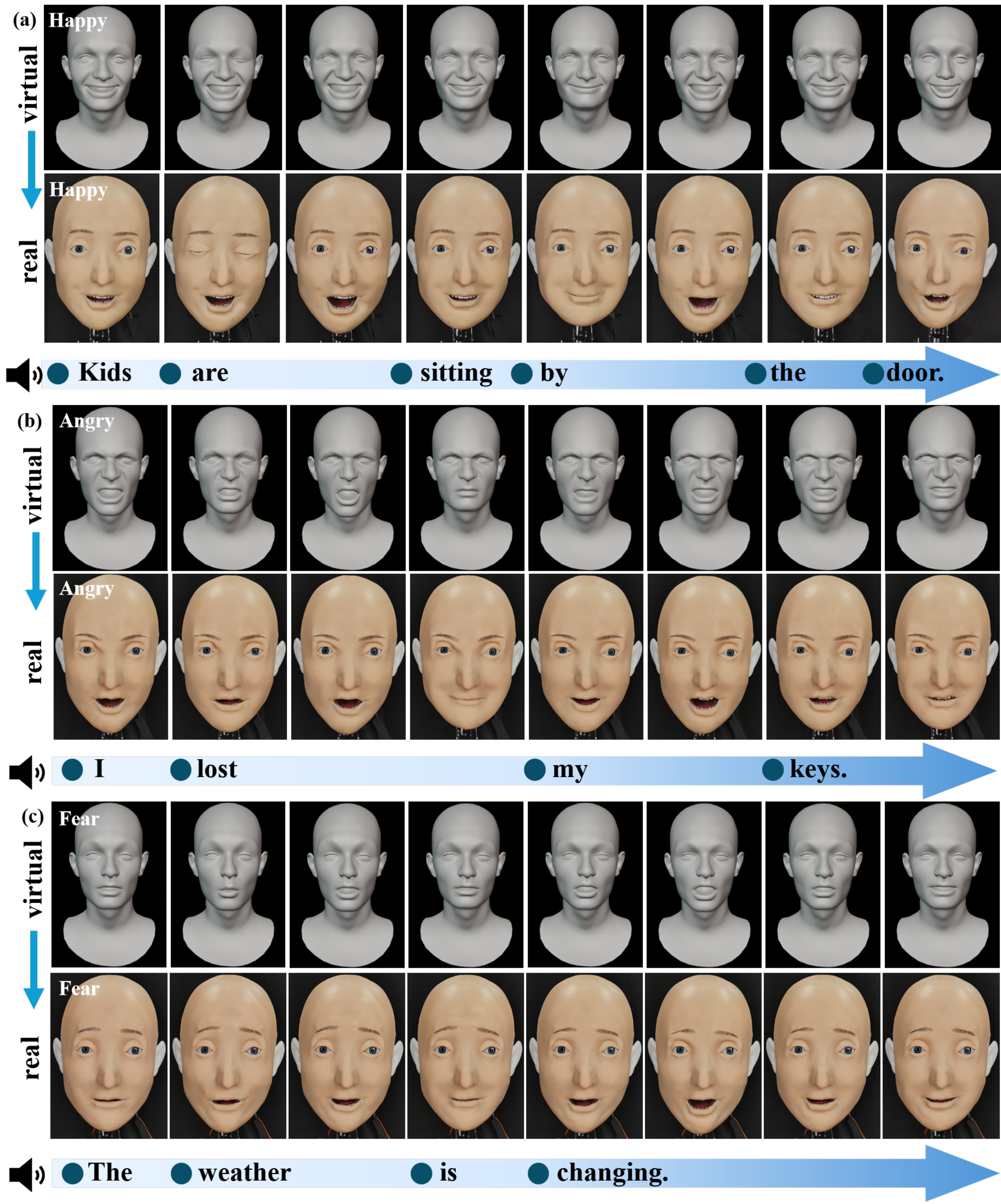}
    \caption{\small \textbf{Complete sentences with diverse facial expressions. }The first row illustrates the virtual expressions generated by our algorithm rendered in Blender, while the second row displays the corresponding
real-world expressions reproduced by the animatronic face. (a) \textbf{Happy} expression while saying "Kids are sitting by the door." 
 (b) \textbf{Angry} expression while saying "I lost my keys." (c) \textbf{Fear} expression while saying "The weather is changing." }
    \label{fig:Qualitative_results}
\end{figure*}

In this experiment, we demonstrate the capability of our system to generate diverse facial expressions while speaking complete sentences. Fig.~\ref{fig:Qualitative_results} shows the results of our system where three different expressions — happy, angry and fear — are exhibited while speaking specific sentences. 

In terms of eyebrow movements, each expression exhibits distinct changes that contribute to the emotional tone. For instance, during the happy expression (Fig.~\ref{fig:Qualitative_results}(a)), the eyebrows are raised, conveying joy, while in the angry expression (Fig.~\ref{fig:Qualitative_results}(b)), the brows furrow tightly, indicating intensity and frustration. In the fear expression (Fig.~\ref{fig:Qualitative_results}(c)), the eyebrows are lowered, further reinforcing the emotional state. Similarly, mouth movements play a crucial role in differentiating the expressions. The happy expression (Fig.~\ref{fig:Qualitative_results}(a)) is marked by the corners of the mouth being raised, resembling a smile. Notably, the mouth movements also align with the natural shape of the mouth for speech, closely matching the typical phonemes and enhancing the authenticity of the emotional expression.

The first row shows the virtual expressions generated by our Emotional Facial Animation algorithm, rendered in Blender, and it is clear that our Emotional Facial Animation algorithm is effective in generating diverse expressions during speech. The second row shows the real-world expressions generated by the animatronic face using our inverse learning algorithm, successfully replicating the virtual expressions and demonstrating the system’s ability to bridge the gap between virtual generation and real-world facial actuation, resulting in realistic and expressive emotional responses.

\begin{figure}
    \centering
    \includegraphics[width=1.0\linewidth]{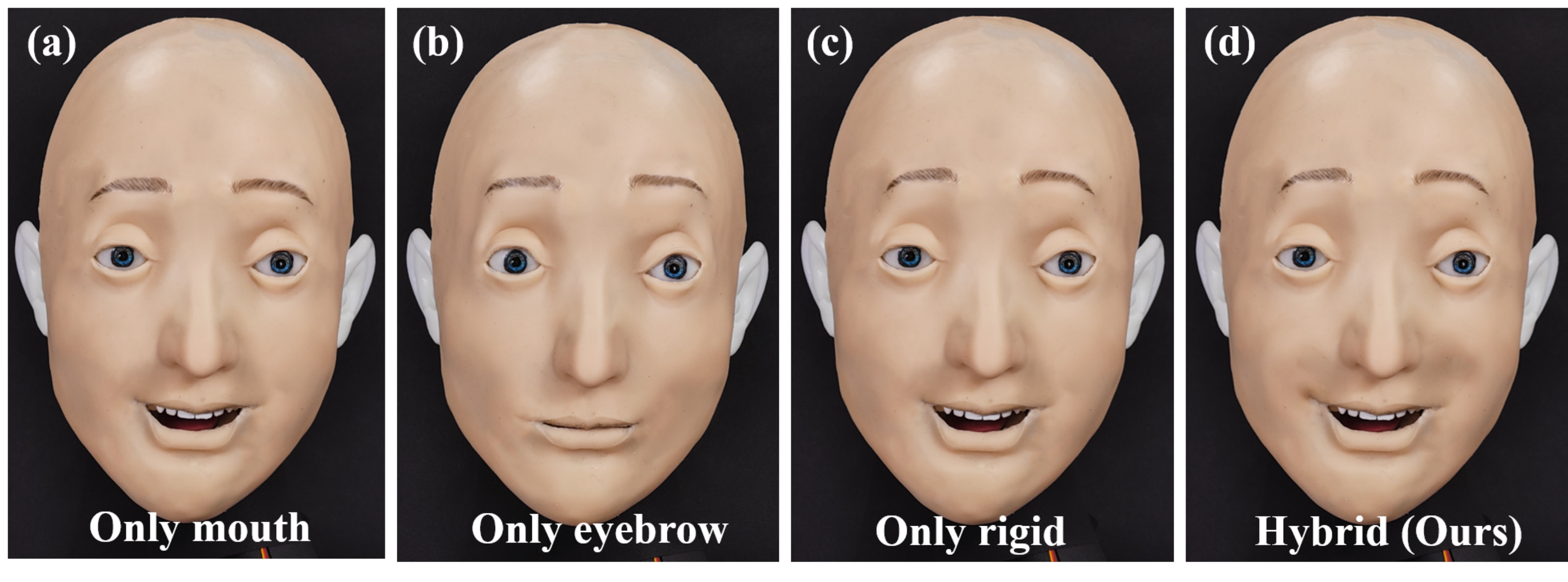}
    \caption{\small \textbf{Comparison of happy expression using different actuation systems. } (a) Mouth-Only System. (b) Eyebrow-Only System. (c) Rigid-Driven System w/o tendon driven. (d) Hybrid-Driven System (Ours). }
    \vspace{-10pt}
    \label{fig:hardware_contrast}
\end{figure}

\subsection{Qualitative Results}
\label{4.4}

To evaluate the overall effectiveness of our mechanical face platform in generating realistic facial expressions, we utilize a video-based facial expression recognition network ~\cite{liu2023expression}. This network is used to classify the generated facial expressions—happy, angry, disgust, and fear—based on the videos produced by our platform. The classification results are represented using a confusion matrix, which provides a detailed breakdown of the system's performance for each expression type.

In the experiment, we randomly select 100 sentences, with each paired with four basic expressions, resulting in a total of 400 videos, each lasting approximately 2 seconds. Table~\ref{tab:confusion_matrix} presents the performance of the full system. The matrix shows high accuracies for happy (90\%) and angry (91\%) expressions, reflecting its strong ability to generate dynamic and expressive movements. While the relatively lower accuracies for disgust (66\%) and fear (73\%) indicate that these expressions are more easily confused, suggesting less distinct differences between them. 

To evaluate lip synchronization, we use the Lip Vertex Error (LVE), as applied in MeshTalk~\cite{richard2021meshtalk} and FaceFormer~\cite{fan2022faceformer}. LVE measures the maximum $\ell_2$ error among all lip vertices for a given frame. Specifically, for ground truth and predicted lip vertices \( \boldsymbol{L}_{\text{gt}} \) and \( \boldsymbol{L}_{\text{pred}} \), LVE is defined as:

\begin{equation}
\text{LVE} = \max_i \| \boldsymbol{L}_{\text{gt}, i} - \boldsymbol{L}_{\text{pred}, i} \|_2.
\end{equation}

However, LVE alone doesn't capture the full emotional expression, particularly in regions like the eyes and forehead. To address this, we use the Emotional Vertex Error (EVE), which calculates the maximum $\ell_2$ error of the vertex displacement in these regions. For vertices in the eye and forehead areas, the EVE is given by:

\begin{equation}
\text{EVE} = \max_j \| \boldsymbol{V}_{\text{gt}, j} - \boldsymbol{V}_{\text{pred}, j} \|_2,
\end{equation}
where \( \boldsymbol{V}_{\text{gt}} \) and \( \boldsymbol{V}_{\text{pred}} \) are the ground truth and predicted coordinates for the selected regions.

The evaluation results in Table~\ref{table:3detf} show that our method outperforms existing approaches in both LVE and EVE on the RAVDESS~\cite{livingstone2018ryerson} and HDTF~\cite{zhang2021flow} datasets. Our model achieves the lowest LVE and EVE scores on both emotional and neutral speech, indicating superior lip synchronization and emotional expressiveness. On the RAVDESS~\cite{livingstone2018ryerson} dataset, we outperform EmoTalk with an LVE of 2.426 mm and an EVE of 2.389 mm. Similarly, on the HDTF~\cite{zhang2021flow} dataset, we achieve an LVE of 2.491 mm and an EVE of 2.252 mm, surpassing other methods like MeshTalk~\cite{richard2021meshtalk} and FaceFormer~\cite{fan2022faceformer}. By introducing the Dynamic Time Warping (DTW) algorithm, we achieve better audio feature alignment, which leads to more accurate lip synchronization and consequently a lower LVE. 
\vspace{-5pt}

\begin{table}[h!]
\centering
\caption{\small Confusion Matrix of recognizing four basic expressions}
\begin{tabular}{c|cccc}
    \toprule
\textbf{True \textbackslash Predicted} & Happy (\%) & Angry (\%) & Disgust (\%) & Fear (\%)\\ 
\midrule
Happy (\%) & \textbf{90} & 2  & 3  & 5  \\
Angry (\%) & 3  & \textbf{91} & 4  & 2  \\
Disgust (\%) & 6  & 3  & \textbf{66} & 25 \\
Fear (\%) & 5  & 2  & 20 & \textbf{73} \\ 
    \bottomrule
\end{tabular}
\label{tab:confusion_matrix}
\end{table}

\begin{table}[]

\caption{\small Quantitative evaluation results on RAVDESS~\cite{livingstone2018ryerson} and HDTF datasets (3D-ETF dataset)~\cite{zhang2021flow}}
\resizebox{\linewidth}{!}{
\begin{tabular}{@{}ccccc@{}}
\toprule
 & \multicolumn{2}{c}{RAVDESS~\cite{livingstone2018ryerson} (emotion)} 
 & \multicolumn{2}{c}{HDTF~\cite{zhang2021flow} (no emotion)}                         \\ \midrule
Method        & \begin{tabular}[c]{@{}c@{}}LVE (mm)$\downarrow$ \end{tabular} & \begin{tabular}[c]{@{}c@{}}EVE (mm)$\downarrow$\\ \end{tabular} & \begin{tabular}[c]{@{}c@{}}LVE (mm)$\downarrow$\\ \end{tabular} & \begin{tabular}[c]{@{}c@{}}EVE (mm)$\downarrow$ \end{tabular} \\ \hline
VOCA~\cite{cudeiro2019capture}  & 5.091    & 4.188     & 4.447   & 3.286  \\
MeshTalk~\cite{richard2021meshtalk}  & 3.459    & 3.386   & 3.886  & 3.124 \\
FaceFormer~\cite{fan2022faceformer}    & 3.247  & 3.757   & 3.374  & 3.142  \\
EmoTalk~\cite{peng2023emotalk}         & 2.762  & 2.493  & 2.892  & 2.364  \\
\textbf{Ours} & \textbf{2.426}   & \textbf{2.389}  & \textbf{2.491}  & \textbf{2.252}  \\\bottomrule
\end{tabular}}
\label{table:3detf}
\end{table}

\subsection{Analysis of Hybrid Actuation Hardware Design}

Fig.~\ref{fig:hardware_contrast} compares the effectiveness of different actuation systems in generating a happy expression. Fig.~\ref{fig:hardware_contrast}(a) shows the result with only the mouth moving, which leads to a limited expressiveness. Fig.~\ref{fig:hardware_contrast}(b) depicts the result with only the eyebrow moving, which fails to capture the full range of facial emotions. Fig.~\ref{fig:hardware_contrast}(c) demonstrates the outcome using a rigid actuation system, which offers better control but still lacks subtle micro-expressions. In contrast, Fig.~\ref{fig:hardware_contrast}(d) presents the hybrid actuation system (ours), which combines both rigid and tendon-driven mechanisms. This hardware design produces the most dynamic and expressive facial motion, highlighting its superiority in conveying emotional expressions.

To quantitatively assess the effectiveness of the hybrid actuation system, we use the same method mentioned in Sec.~\ref{4.4}. The classification accuracy is measured using accuracy across four configurations mentioned above. Following the same experimental setup as the main experiment, 400 videos are generated for each configuration and evaluated to compare performance. Additionally, we use landmark distance to measure the discrepancy between the mechanical face's expressions and the virtual expressions. We observe a larger facial landmark displacement in subtle areas between the rigid-only system and ours (as shown in Table~\ref{tab:landmark_comparison}), confirming the hybrid system’s superior expressiveness. For clarity, all landmark distances are converted from pixel units to millimeters using camera-to-face calibration based on known facial dimensions of the robot platform.

\begin{table}[h!]
\centering
\caption{\small Comparison of Facial Landmark Displacement}
\renewcommand{\arraystretch}{1.2}
\resizebox{1\linewidth}{!}{
\begin{tabular}{cccc}
\toprule
Subtle Area & Rigid-Only System (mm) $\downarrow$ & Hybrid System (mm) $\downarrow$\\
\midrule
Cheek              & 2.10                   & \textbf{0.95}               \\
Nose               & 1.18                   & \textbf{0.26}              \\
\bottomrule
\end{tabular}%
}
\label{tab:landmark_comparison}
\end{table}

\vspace{-10pt}

\begin{table}[h!]
\centering
\caption{\small Classification Accuracy for Videos Generated by Different Hardware Configurations}
\renewcommand{\arraystretch}{1.2}
\resizebox{1\linewidth}{!}{
\begin{tabular}{ccccccc}
\toprule
Hardware Configuration & Happy (\%) & Angry (\%) & Disgust (\%) & Fear (\%) & Mean (\%) \\
\midrule
Mouth-Only System        & 65 & 41 & 31 & 35 & 43 \\
Eyebrow-Only System        & 47 & 50 & 39 & 43 & 44.75 \\
Rigid-Driven System w/o tendon driven     & 72 & 75 & 51 & 60 & 64.5 \\
Hybrid-Driven System (Ours)    & \textbf{90} & \textbf{91} & \textbf{66} & \textbf{73} & \textbf{80} \\
\bottomrule
\end{tabular}%
}
\label{tab:ablation_hardware}
\end{table}
The results, summarized in Table~\ref{tab:ablation_hardware}, indicate that the hybrid-driven system significantly outperforms other configurations across all metrics. The addition of the tendon-driven module significantly improves the mAP from 64.5\% to 80\%, with notable gains in happy and angry expressions, increasing from 72\% to 90\% and 75\% to 91\%, respectively, highlighting the crucial contribution of the tendon-driven module to these expressions. In contrast, fear and disgust have slightly lower accuracies, reflecting their complexity. Additionally, the overall performance of the eyebrow-only system surpasses that of the mouth-only system, emphasizing the greater importance of eyebrow movements in facial expressions.
\section{Limitations}

In terms of hardware,  the facial skin and actuators are attached using adhesive, which complicates debugging and adjustments. A more efficient solution could involve fixed attachment points with magnets or clasps. Our experiments also show that the material and thickness of the facial skin impact expression quality, emphasizing the need for further testing to select the best materials and thickness to better replicate natural human facial anatomy. 

On the software side, the current facial expression generation system takes emotional speech as input and produces corresponding facial expressions, which results in relatively high inference times. To enable more natural and real-time interactions between humans and robotics, future work will need to refine the network architecture, focusing on achieving near real-time dialogue capabilities between the human and the robotic.
\section{Conclusion}

In this work, we present \textit{Morpheus}, an animatronic face system that combines hybrid actuation, self-modeling, and speech-driven expression generation for lifelike facial movements. Experimental results show that the hybrid actuation system outperforms traditional rigid or tendon-driven approaches in producing expressive facial movements. The self-modeling network accurately links motor control to facial deformation, enabling smoother expressions. Our speech-driven system produces dynamic and natural facial animations, improving human-robot interaction. We believe that \textit{Morpheus} sets a new benchmark for expressive robotic faces, bringing us closer to realistic, immersive interactions.
{
    \small
    \bibliographystyle{plainnat}
   \bibliography{main}
}
\newpage

In this supplementary material, we provide more details about \textit{Morpheus}, including:

\noindent(1) Emotional Facial Animation Module Training: Details about the training process used for generating emotional facial animations.

\noindent(2) Description of the Mechanical Face Hardware Platform: Details about the hybrid-driven mechanical face platform.

\noindent(3) Face Inverse Learning Module Training: A description of the training approach for the Face Inverse Learning Module.

\noindent(4) Additional Visualization: Additional visualizations showcasing the system’s performance in generating diverse facial expressions with speech.

\noindent(5) Generalizability of the Self-Modeling Network: Details about how the self-modeling network generalizes across different robotic platforms with varying actuator configurations.

\noindent(6) User Perception Study: a user study evaluating Morpheus' facial expressions in comparison to real human faces and other robotic systems.

\noindent(7) Hardware Noise Discussion: the strategies used to minimize noise in the mechanical face system.

\noindent(8) Material Selection and Expressiveness: Details about the selection of materials and their impact on the expressiveness and realism of the mechanical face.

\noindent(9) Latency in Real-Time Human-Robot Interaction: Details about the system’s real-time performance.

\begin{figure*}
    \centering
    \includegraphics[width=1.0\linewidth]{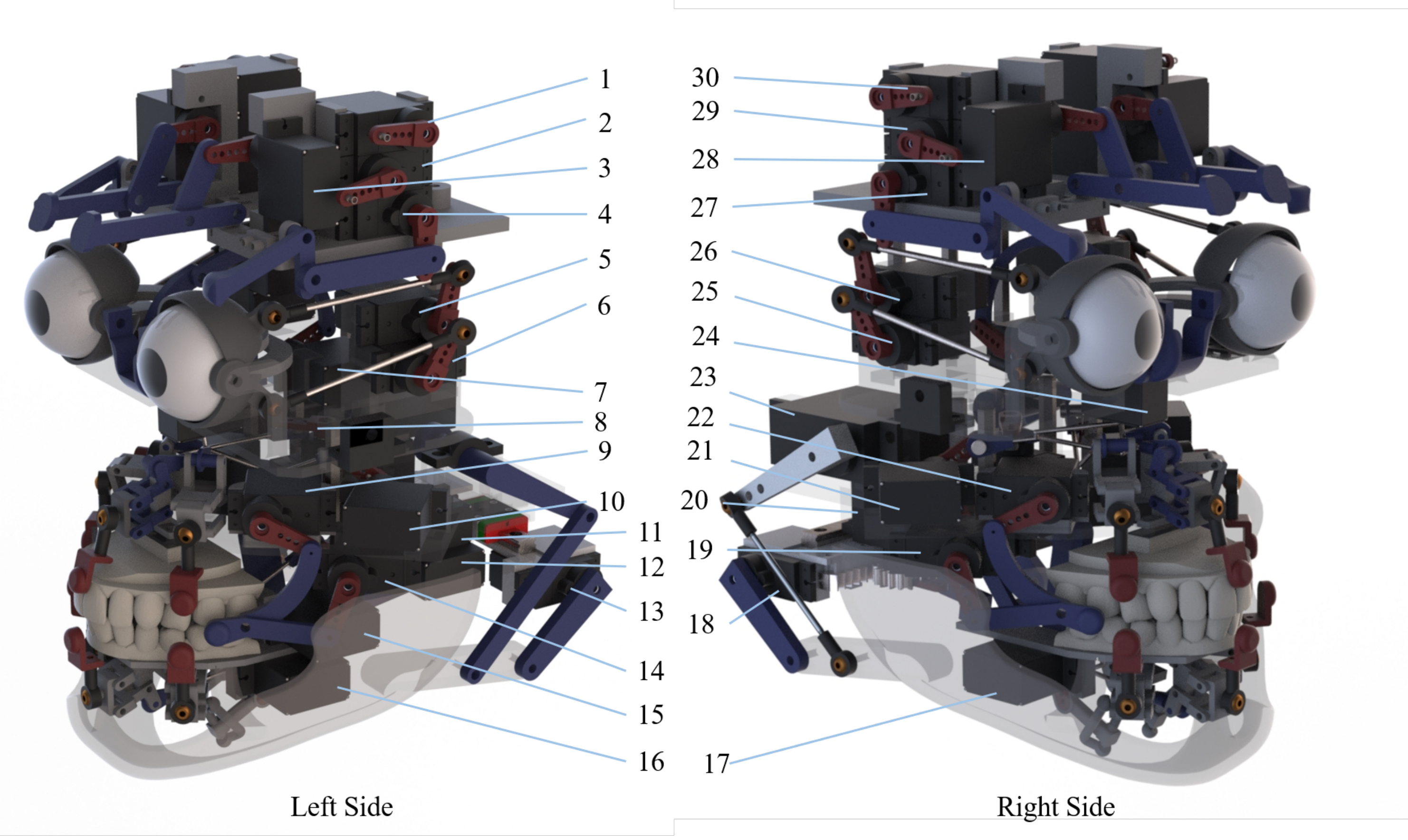}
    \caption{\textbf{Motor Distribution and Control Areas of the Mechanical Face Hardware Platform.} 1 - Left Cheek (tendon drive). 2 - Left Nose (tendon drive). 3 - Left Eyebrow Center. 4 - Left Eyebrow Peak. 5 - Left Upper Eyelid. 6 - Left Lower Eyelid. 7 - Eyes Up/Down Gaze. 8 - Upper Lip Center. 9 - Left Upper Corner Mouth Movement. 10 - Upper Lip Right. 11 - Tongue Extension/Retraction. 12 - Tongue Up/Down. 13 - Chin (passive). 14 - Left Lower Corner Mouth Movement. 15 - Lower Lip Center. 16 - Lower Lip Left. 17 - Lower Lip Right. 18 - Chin (active). 19 - Right Lower Corner Mouth Movement. 20 - Chin Left/Right. 21 - Upper Lip Left. 22 - Right Upper Corner Mouth Movement. 23 - Mouth Opening/Closing. 24 - Eyes Left/Right Gaze. 25 - Right Lower Eyelid. 26 - Right Upper Eyelid.  27 - Right Eyebrow Peak. 28 - Right Eyebrow Center. 29 - Right Nose (tendon drive). 30 - Right Cheek (tendon drive). }
    \label{fig:sup_motor}
\end{figure*}

\subsection{Emotional Facial Animation Module Training}
For training the \textit{Emotional Facial Animation} module, preprocessed video and audio data from the dataset are used as inputs. The video is processed to a consistent rate of 30 frames per second (fps), and the audio is sampled at 16 kHz. Each video frame is represented by 33 blendshape coefficients. 

The model is trained in an end-to-end fashion with the Adam optimizer, where the learning rate is set to 1e-4 and the batch size is 16. Training is performed on a single NVIDIA A100 GPU, and the entire process requires around 4 hours (80 epochs).

\subsection{Description of the Mechanical Face Hardware Platform}

Our mechanical face hardware platform is composed of 33 actuators, each carefully placed to control various facial regions and enable complex expressions. Fig.~\ref{fig:sup_motor} provides a detailed layout of the actuator distribution, showcasing the strategic placement of motors for maximum expressiveness and functionality. The remaining three motors, labeled 31-33, control the neck’s nodding, shaking, and rotation movements, which are not marked in the figure.

Of the 33 motors, 26 are dedicated to controlling the movements of key facial areas: the eyebrows, eyes, nose, cheeks, mouth, and chin. These motors are numbered 1-6, 8-10, 13-23, and 24-30, and their precise placement allows for nuanced, lifelike expressions that are crucial for emotional communication.

\subsection{Face Inverse Learning Module Training}
As discussed in Sec. IV-A, the dataset used to train the MLP network consists of control commands for 26 actuators that control various facial expressions. These control commands are randomly generated to simulate diverse facial movements. To prevent structural interference during actuator motion, we have thoroughly tested and defined the control command range for each actuator (refer to Tab.~\ref{tab:motor_control}). Without constraints, the control commands for each actuator span from 0 to 2000. Each motor's control is normalized to the range \([0, 1]\). Then we train the MLP network using the $\ell_1$ loss. The overall process takes 10-15 minutes on a single NVIDIA 3090 GPU with a  1e-5 learning rate.
\subsection{Visualization of Virtual and Real Mechanical Faces with Emotions Speaking Specific Sentences}

\begin{figure*}
    \centering
    \includegraphics[width=0.9\linewidth]{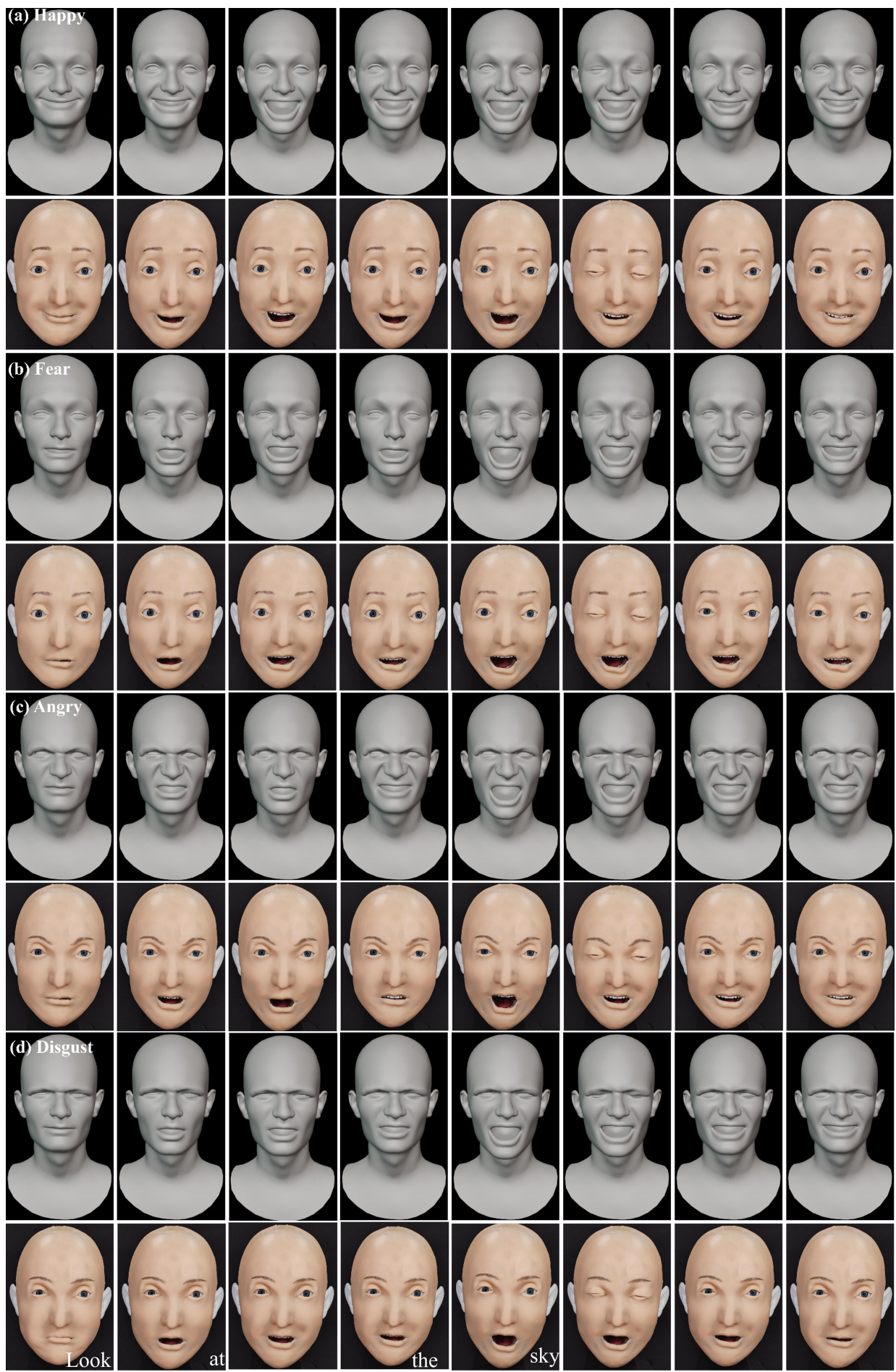}
    \caption{\small Visualization of \textit{Morpheus} interpreting the sentence "look at the sky" with happy, fear, angry and disgust expressions, respectively.}
    \label{fig:four_expressions}
\end{figure*}

Fig.~\ref{fig:four_expressions} illustrates a sequence of images generated by our algorithm and mechanical face platform, where four basic expressions—happy, fear, angry, and disgust—are used to say the sentence "look at the sky." The comparison between (a) to (d) clearly shows the facial variations corresponding to different expressions. The vertical comparison in each subfigure demonstrates the consistency between between the virtual facial expressions and the real facial expressions, with the \textit{Face Inverse Learning} module mapping the virtual expressions to the real ones.

\begin{table*}
\centering
\caption{Motor Control Parameter Range}
\renewcommand{\arraystretch}{1.2}
\resizebox{1\linewidth}{!}{
\begin{tabular}{cccccc}
\toprule
\textbf{Motor ID} & \textbf{Control Position} & \textbf{Lower Bound} & \textbf{Start Value} & \textbf{Upper Bound} \\
\midrule
1 & Left Cheek & 1800 & 1800 & 1400 \\
2 & Left Nose  & 800 & 800 & 0 \\
3 & Left Eyebrow Center & 980 & 1080 & 1480 \\
4 & Left Eyebrow Peak & 1100 & 800 & 500\\
5 & Left Upper Eyelid & 1700 & 1300 & 600\\
6 & Left Lower Eyelid &1000 &1200 & 1800\\
7 & Eyes Up/Down Gaze & 300 & 1000 & 1500\\
8 & Upper Lip Center &1000 &1000 &600\\
9 & Left Upper Corner Mouth Movement &500 &1000 &1500\\
10 & Upper Lip Right & 1000 & 1000 & 1200\\
11 & Tongue Extension/Retraction & 800 & 1000 & 1200\\
12 & Tongue Up/Down & 800 & 500 & 300\\
13 & Chin (passive) & 1200 &1600 & 1600\\
14 & Left Lower Corner Mouth Movement & 1000 & 1000 & 1500\\
15 & Lower Lip Center & 1300 & 1300 & 1900\\
16 & Lower Lip Left & 1500 & 1500 & 800 \\
17 & Lower Lip Right & 800 & 800 & 1200\\
18 & Chin (active) & 1200 & 1600 & 1600\\
19 & Right Lower Corner Mouth Movement & 1000 & 1000 & 500\\
20 & Jaw Left/Right & 400 & 1000 & 1600\\
21 & Upper Lip Left & 1000 & 1000 & 800\\
22 & Right Upper Corner Mouth Movement & 1500 & 1000 & 500\\
23 & Mouth Opening/Closing & 1150 & 1050 & 1050\\
24 & Eyes Left/Right Gaze & 800 & 1150 & 1500\\
25 & Right Lower Eyelid & 1200 & 1000 & 400\\
26 & Right Upper Eyelid & 500 & 1000 & 1600\\
27 & Right Eyebrow Peak & 900 & 1200 & 1500\\
28 & Right Eyebrow Center & 1300 & 1200 & 800\\
29 & Right Nose & 1200 & 1200 & 2000\\
30 & Right Cheek & 0 & 0 & 600\\
31 & Neck Rotation & 1500 & 1750 & 2000\\
32 & Neck Nodding/Shaking (left) & 900 & 1200 & 1300\\
33 & Neck Nodding/Shaking (right) & 1200 & 900 & 800\\
\bottomrule
\end{tabular}%
}
\label{tab:motor_control}
\end{table*}

\subsection{Generalizability of the Self-Modeling Network.} 
\begin{figure}[]
    \includegraphics[width=1.0\linewidth]{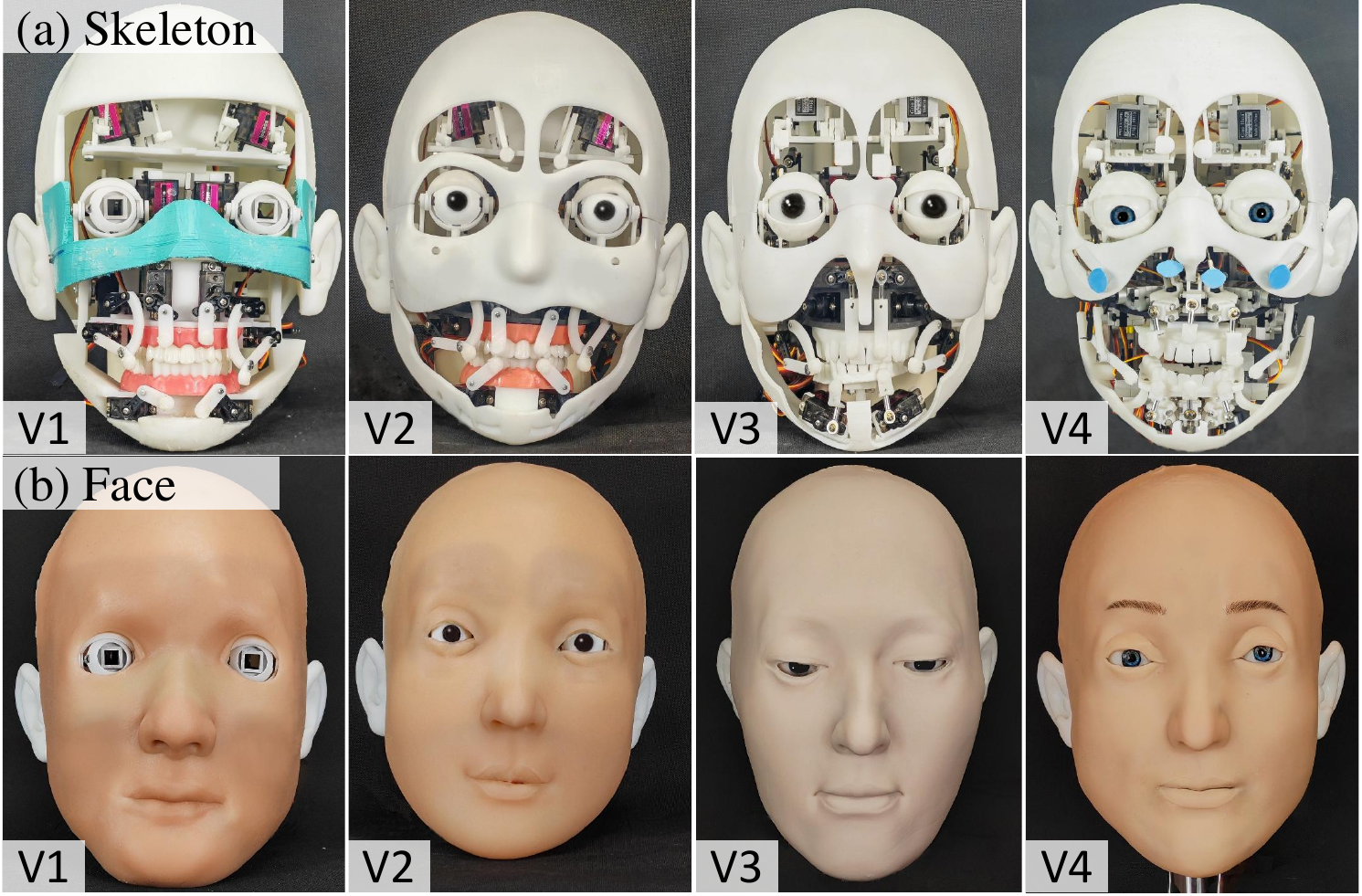}

    \caption{\textbf{Morpheus design evolution across four iterations.} (a) Internal skeletal structures. (b) External facial appearances with silicone skin applied.}
    \label{fig:face_version}

\end{figure}

Our self-modeling framework has been successfully deployed across multiple robotic platforms with varying actuator configurations, including four versions of Morpheus with 17, 16, 23, and 26 degrees of freedom (DoF) for facial expression control (see Fig.~\ref{fig:face_version}). For each version, we adapted the final layer of the MLP to align with the specific number of actuators, while preserving the core architecture and its ability to connect blendshape coefficients with motor control signals. We evaluated the network's performance using the Mean Absolute Error (MAE) between predicted facial landmarks and the ground truth. 

The results demonstrate the self-modeling network performed well across all four versions, showing its ability to adapt to different actuator configurations and improve facial expression accuracy. As shown in Table~\ref{tab:landmark_distances}, the network's generalizability is confirmed by the varying landmark distances across the different versions.

\begin{table}[h!]
\centering
\caption{\small Landmark Distances for Different Versions of the Self-Modeling Network}
\renewcommand{\arraystretch}{1.2}
\resizebox{1\linewidth}{!}{
\begin{tabular}{ccccc}
\toprule
Version & V1 & V2 & V3 & V4 \\
\midrule
Landmark Distance (mm) & 1.652  & 1.313 & 1.206 & 0.748  \\
\bottomrule
\end{tabular}%
}
\label{tab:landmark_distances}
\end{table}

\subsection{User Perception Study} 
A user study with 50 participants (age 15-50, balanced gender) evaluated Morpheus' expressions compared to real human faces and other robotic systems. The study had three parts: \textbf{1) Static Expression Comparison}, where participants selected the most realistic expressions from Morpheus and three other recent animatronic faces [9, 19, 32]
(sourced from their published papers and demos); \textbf{2) Speech-Driven Comparison}, where participants compared synchronized speech-driven videos of Morpheus and [32] (from public demo) for expressiveness and selected the better one; \textbf{3) Realism Scoring}, where participants rated 20 Morpheus speech-emotion clips against human recordings on a 1-10 Likert scale for naturalness and emotional clarity. Results showed that Morpheus was preferred in 84\% of cases for realistic emotions, with a mean rank of 1.18. In speech-driven comparisons, 96\% of participants selected Morpheus as more expressive than [32]. For realism scoring, Morpheus averaged 7.6 on a 1-10 Likert scale, confirming its effectiveness in generating lifelike expressions with hybrid actuation and neural-driven control.

\begin{figure}[]
    \includegraphics[width=1.0\linewidth]{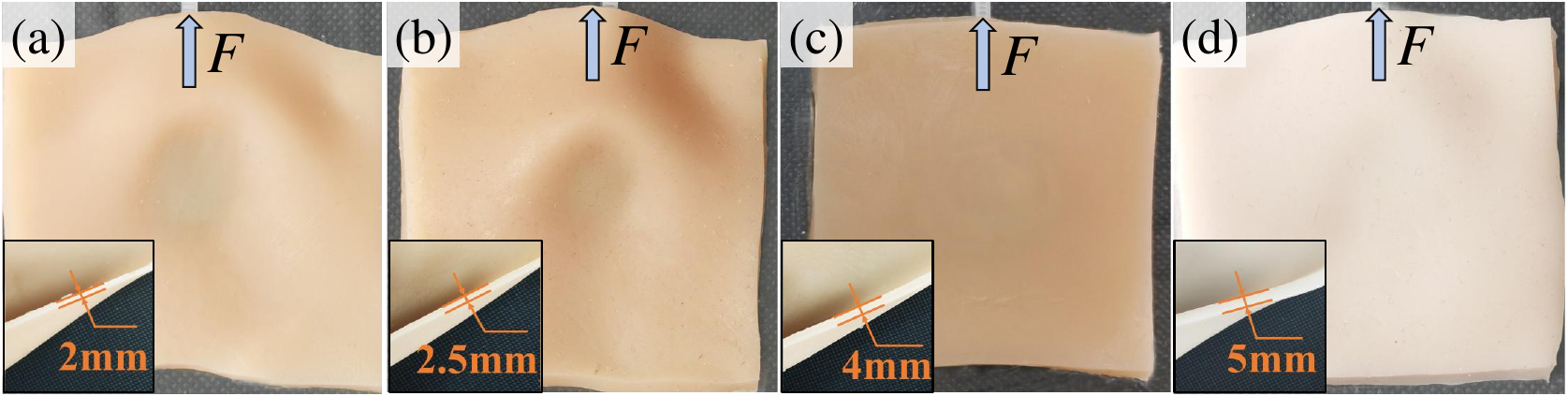}

    \caption{\textbf{Impact of silicone skin thickness on tendon-driven facial deformation.} (a-b) For thinner skins (2mm and 2.5mm), the deformation is unnaturally localized around the pasted region (indicated by the base of the arrow). (c) At 4mm thickness (our selected configuration), the skin exhibits smooth and natural movement, enabling expressive actuation.  (d) A 5mm thickness is overly rigid, significantly limiting visible deformation.}
    \label{fig:material_thickness}

\end{figure}

\subsection{Hardware Noise Discussion}
To minimize noise, we optimized servo selection and implemented soundproofing measures. We primarily used GUOHUA 9g servos, operating at \textbf{45}-\textbf{50} dB, lower than the 50-60 dB of alternatives like MG90S and within the WHO's \textbf{55} dB limit. For larger TD-8035M servos, 2mm butyl rubber soundproof felt reduced noise by \textbf{20} dB. Future improvements include lubricating gears, increasing PWM frequency, or using brushless servos to reduce noise.

\subsection{Material Selection and Expressiveness}
To optimize expression accuracy and realism, especially in tendon-driven regions, we carefully tested silicone materials. We used A00-30 silicone, known for its skin-like softness, and adopted a 3D-printed mold solution for efficient iteration. The silicone was mixed (1:1 ratio), degassed in a vacuum chamber (-0.1 MPa, 5-10 min), poured into the mold, and cured at 25°C for 4-8 hours, followed by a 24-hour resting period to enhance performance.
Material tension was adjusted by thickness (2mm, 2.5mm, 4mm, 5mm), with the eye area thinned to \textbf{2}mm to prevent localized buildup at the eyelid. For tendon-driven sections, experiments determined that \textbf{4}mm skin provided the best balance between visibility and actuation capability (Fig.~\ref{fig:material_thickness}).

\subsection{Latency in Real-Time Human-Robot Interaction} Our system infers at about \textbf{150} FPS on NVIDIA Jetson AGX Xavier, generates virtual expressions at \textbf{30} FPS, and controls servos at \textbf{50} Hz, enabling smooth and synchronized alignment between virtual and real expressions.

\end{document}